\documentclass[lettersize,journal]{IEEEtran}
\usepackage{amsmath,amsfonts}
\usepackage{algorithm}
\usepackage{array}
\usepackage[caption=false,font=normalsize,labelfont=sf,textfont=sf]{subfig}
\usepackage{textcomp}
\usepackage{stfloats}
\usepackage{url}
\usepackage{verbatim}
\usepackage{graphicx}
\usepackage{cite}
\usepackage{comment}
\usepackage{caption}
\usepackage{graphicx}
\usepackage{algorithm}
\usepackage{algpseudocode}
\usepackage{enumitem}
\setlist[enumerate]{itemsep=0mm}
\usepackage{amsmath}
\usepackage{multirow}
\usepackage{xcolor}

\begin{document}

\title{Deep learning model compression using network sensitivity and gradients}

\author{Madhumitha Sakthi\textsuperscript{1}, Niranjan Yadla\textsuperscript{2}, Raj Pawate\textsuperscript{2}
\thanks{1. Electrical and Computer Engineering, The University of Texas at Austin, USA (madhumithasakthi.iyer@utexas.edu), 2. Tensilica IPG, Cadence Design Systems Inc., USA (pawateb@cadence.com)}}



\maketitle

\begin{abstract}
Deep learning model compression is an improving and important field for the edge deployment of deep learning models. Given the increasing size of the models and their corresponding power consumption, it is vital to decrease the model size and compute requirement without a significant drop in the model's performance. In this paper, we present model compression algorithms for both non-retraining and retraining conditions. In the first case where retraining of the model is not feasible due to lack of access to the original data or absence of necessary compute resources while only having access to off-the-shelf models, we propose the Bin \& Quant algorithm for compression of the deep learning models using the sensitivity of the network parameters. This results in 13x compression of the speech command and control model and 7x compression of the DeepSpeech2 models.
In the second case when the models can be retrained and utmost compression is required for the negligible loss in accuracy, we propose our novel gradient-weighted k-means clustering algorithm (GWK). This method uses the gradients in identifying the important weight values in a given cluster and nudges the centroid towards those values, thereby giving importance to sensitive weights. Our method effectively combines product quantization with the EWGS \cite{EWGS} algorithm for sub-1-bit representation of the quantized models. We test our GWK algorithm on the CIFAR10 dataset across a range of models such as ResNet20, ResNet56, MobileNetv2 and show 35x compression on quantized models for less than 2\% absolute loss in accuracy compared to the floating-point models. 

\end{abstract}


\section{Introduction}
Deep learning models are applied to achieve state-of-the-art results across applications in various fields. The initial success of deep learning models was attributed to the extremely large model parameters and hence their ability to model complex problems. However, given the applicability of deep learning across various fields and their upcoming edge device deployment, it is crucial to reduce the model size and computational complexity without compromising on the model performance. Therefore, recent research has focused on producing lightweight deep learning models, pruning, quantization and clustering techniques for compression. Most often, these methods either focus on storage compression or model quantization for edge computing advantage. Also, these methods need extensive re-training of the compressed model starting from the floating point model which is already heavily trained. While re-training of these models is compute-intensive along with the requirement of the knowledge of the initial hyperparameters used for training the original model, in order to achieve extreme compression for the least loss in accuracy, it is vital to retrain the model. 

\begin{figure}[ht!]
\begin{center}
\includegraphics[width=\linewidth]{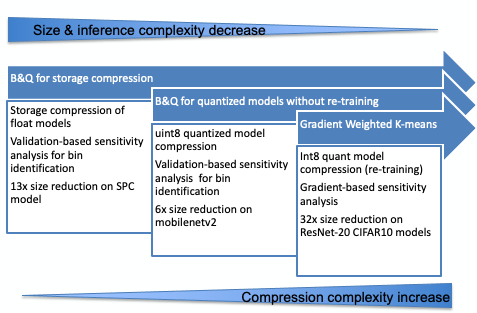}
\centering
\caption{The overall compression algorithm for various models. As the compression complexity increases, the size and inference complexity of the model decreases.}
\label{Algo}
\end{center}
\end{figure}

In this paper, we present 3 such compression cases for varying storage, compression complexity and accuracy requirements. First, our original Bin \& Quant method \cite{icassp} is specifically designed for compressing the pre-trained models without having to retrain them after compression. In this method, we use sensitivity analysis to guide us in identifying the appropriate bins for a particular layer in a model and cluster the bin values to represent them using a single label and hence introduce storage compression. This method was specifically developed for float activation models and for storage compression. 
The main motivation for developing a technique for compression without retraining is attributed to the need for the original training data's availability for retraining. In this era of growing need for privacy and security concerns, often obtaining access to the original data could be difficult\cite{WS}. Also, it is important to have prior knowledge about the hyperparameters used for training the original model and if there is no access to the same amount of compute capacity that was used to train the initial model, there will be a need to explore hyperparameters for retraining the model again. In addition to this, setting up the system for retraining is also an additional task that is time-consuming and hence expensive. Therefore, we propose a compression algorithm without the need for retraining.
In the second case, we extend our Bin \& Quant algorithm to quantized models and apply the algorithm for compression of the uint8 weight values without retraining. This compression scheme works on top of the quantized models and provides extra compression at minimal cost in accuracy and compression complexity.

Finally, in cases where retraining is feasible, we present a gradient-weighted k-means(GWK) algorithm in addition to quantization-aware training of the model using the EWGS \cite{EWGS} algorithm for storage compression of the quantized models. This method is proposed for utmost model compression while taking advantage of quantized weights and activation to retain the edge deployment advantages of the quantized models. Since we use product quantization, similar to \cite{DKM,ATB}, we achieve sub-1-bit representation per weight while the activation is also quantized. Our GWK method utilizes the gradients in identifying the sensitive parameters of the network. We empirically show that the network parameters perturbed based on a higher gradient lead to a drop in accuracy compared to randomly perturbed weights. Thereby, emphasizing the fact that the gradients are capable of identifying sensitive parameters of the network and this information aids in nudging the centroids of the clusters towards these sensitive weights in our GWK algorithm. 
Therefore, we present these three algorithms for three different scenarios. In the first case, where retraining is not feasible and the model has float weights and activations, Bin \& Quant would be used for storage level compression. In the second case, we propose to use Bin \& Quant for the storage compression of the quantized models. Finally, the third algorithm is proposed for the utmost model compression in conditions where retraining is possible and least/no accuracy drop is required. 

We tested our first algorithm on speech recognition models and showed 13x compression in the case of the speech command and control model, and 7x compression of the deepspeech2 model and this included a range of speech commands and large vocabulary speech recognition models and VGG16 network. The second algorithm was tested on computer vision models such as Mobilenetv2 and Resnet since these models are widely used as post training quantized models, especially in edge devices for image classification tasks. In this case, we achieved 6.32x compression of the Mobilenetv2 model with quantized activations. Finally, the third algorithm was tested on ResNet, MobileNet models and we achieved sub-1-bit representation per weight of the models with quantized activations. 
In figure \ref{Algo}, we describe the three compression algorithm and their relevant result. As the compression complexity increases, the size and inference complexity decrease. 

Therefore, the main contribution of this paper is the following:
\begin{itemize}
    \item A novel Bin \& Quant model compression algorithm for off-the-shelf models without retraining the float weight and activation parameters for storage compression. 
    \item The Bin \& Quant algorithm extended to integer quantized models for compression of weight and activation quantized models without retraining. 
    \item Empirically established the significance of gradients in identifying the sensitive parameters of the network. 
    \item A novel gradient weighted k-means algorithm for utmost model compression of the quantized models using EWGS \cite{EWGS}. The gradients nudge the centroids of the cluster towards sensitive parameters and the algorithm achieves a sub-1-bit weight representation of quantized parameters.
\end{itemize}

We organize the rest of the paper as follows. In section two, we give a brief overview of the existing research on deep learning model compression algorithms for storage compression and quantization methods. In section three, we present our three proposed algorithms for retraining and non-retraining-based compression methods. In section four, the experimental evaluation of the proposed algorithms is presented across various models and finally, we conclude our methods in section five.

\begin{figure*}[!h]
\centering
\begin{minipage}[c]{0.25\linewidth}
\includegraphics[width=\linewidth]{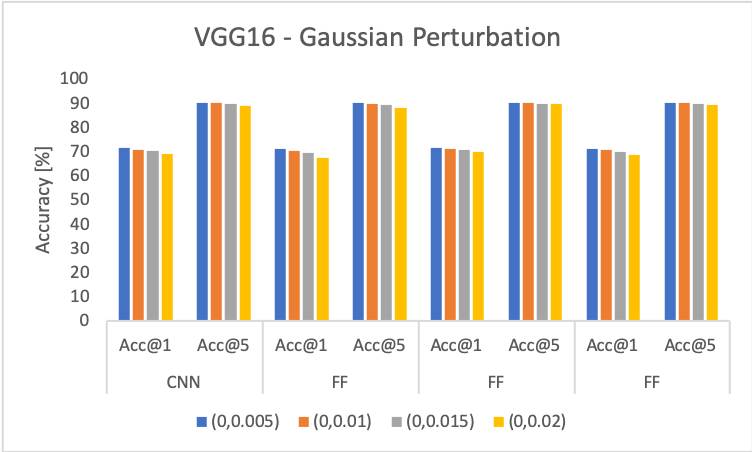}
\end{minipage}\hfill
\begin{minipage}[c]{0.25\linewidth}
\includegraphics[width=\linewidth]{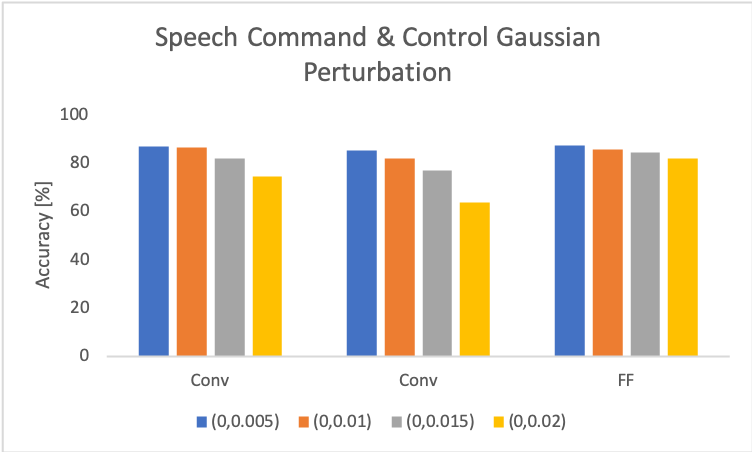}
\end{minipage}\hfill
\begin{minipage}[c]{0.25\linewidth}
\includegraphics[width=\linewidth]{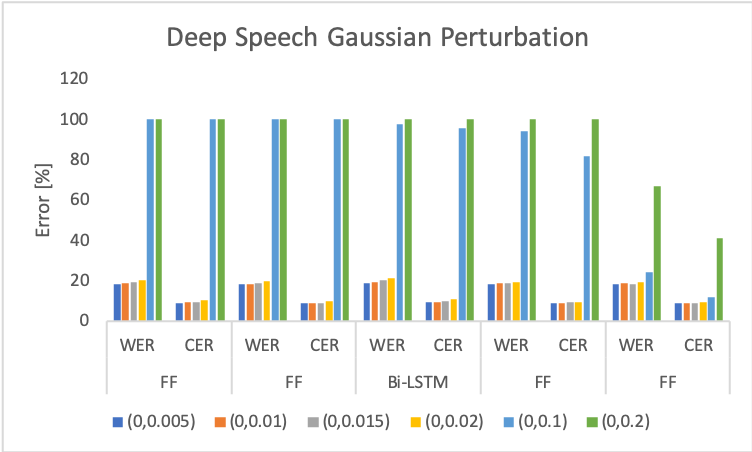}
\end{minipage}\hfill
\begin{minipage}[c]{0.25\linewidth}
\includegraphics[width=\linewidth]{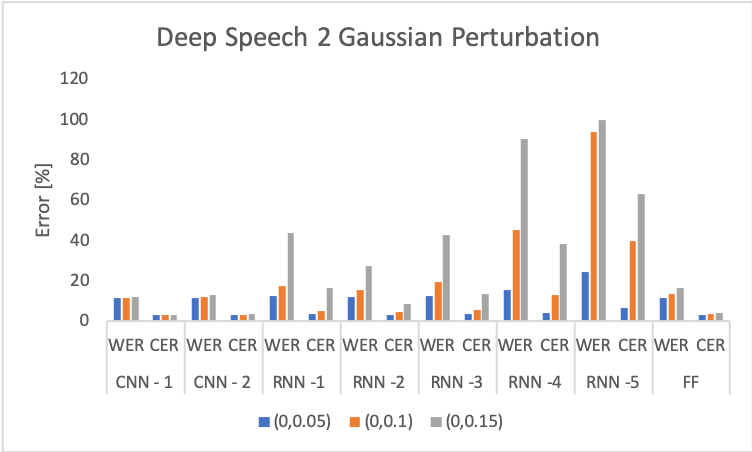}
\end{minipage}\hfill\\[1ex]
%
\begin{minipage}[c]{0.25\linewidth}
\includegraphics[width=\linewidth]{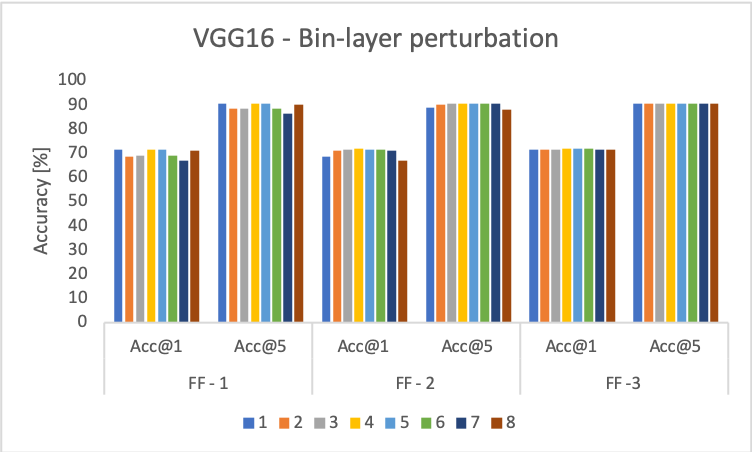}
\end{minipage}\hfill
\begin{minipage}[c]{0.25\linewidth}
\includegraphics[width=\linewidth]{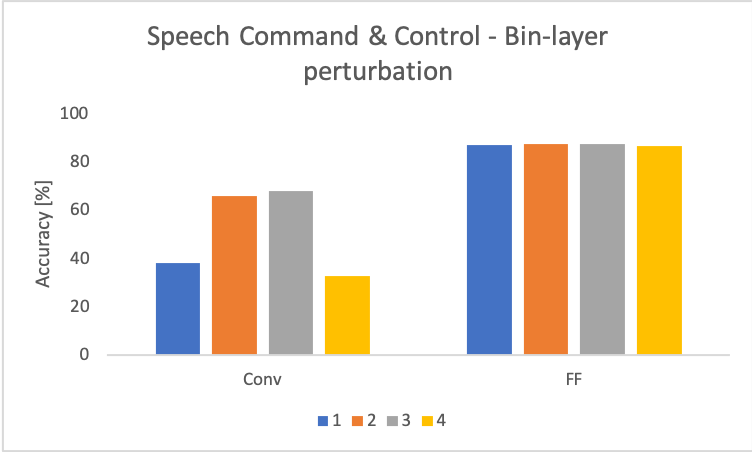}
\end{minipage}\hfill
\begin{minipage}[c]{0.25\linewidth}
\includegraphics[width=\linewidth]{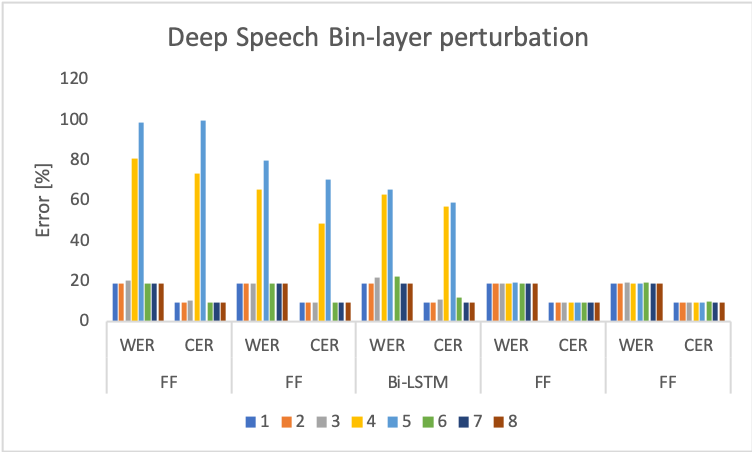}
\end{minipage}\hfill
\begin{minipage}[c]{0.25\linewidth}
\includegraphics[width=\linewidth]{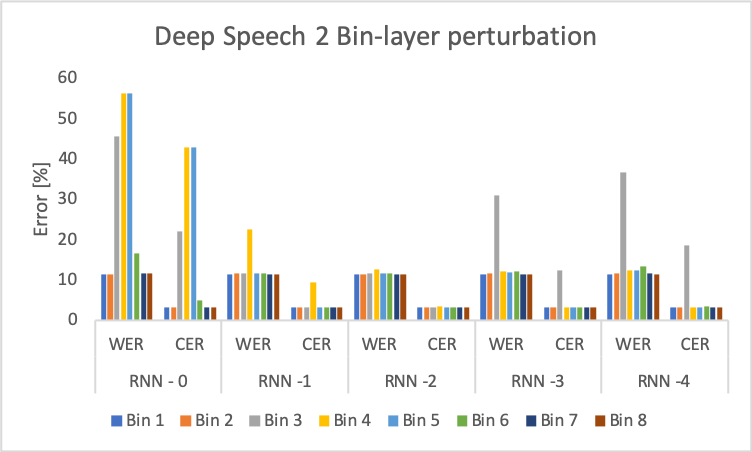}
\end{minipage}

\caption{From left to right, the effect of Gaussian perturbation is shown in the first row, and the effect of a $\pm 50\%$ relative magnitude perturbation is shown in the second row for VGG16, SPC, measured using accuracy and  DS, DS 2 measured using error rate.}
\label{GaussBin-layerPerturb}
\end{figure*}

\section{Related Work}
Clustering-based model compression techniques have recently gained a lot of attention due to their ability to represent multiple weight values using a single label and hence save on storage space. Particularly, the work by \cite{DC} has been widely used and developed further. In \cite{DC}, the authors presented a multi-stage compression pipeline of clustering, re-training and Huffman encoding of various vision models such as VGG16, AlexNet. Recently, the work by \cite{ATB} extended this clustering technique to use product quantization in addition so that multiple weight values in proximity will be represented by a single label value and thereby leading to the ability to represent with sub-1-bit per weight value. Their k-means algorithm is optimized to preserve the output activation rather than the weights directly and showed 29x compression in the case of ResNet-18 and 18x compression in the case of ResNet-50 model. In the \cite{DKM} paper, the non-differentiable k-means clustering technique was addressed by adding attention matrices to guide cluster formation and enabled joint optimization of DNN parameters and cluster centroids. They could compress the Mobilenet-v2 to 3 bits/weight and ResNet-18 to 0.717 bits/weight. However, in all of the above methods, the authors focused on weight storage compression and retained float activations.
In DKM \cite{DKM} the authors propose to use attention matrices to guide centroid formation and therefore, in the case of a large ResNet-50 model with 8/8 i.e., 8-bit clusters and 8-d subvector size, the model does not converge due to out-of-memory error. However, in our proposed GWK, we use a 1-d gradient vector, the same size as that of our original model weight that is being compressed without the need for an extra attention matrix parameter to guide cluster formation. Also, while DKM only focuses on storage compression, we provide results on both storage compression and weight quantization. Finally, in DKM the attention matrices are derived based on the distance metric and this is used to guide the centroid formation. Whereas, we use the gradients to guide us in the centroid formation motivated by the empirically tested hypothesis that gradients are capable of identifying the sensitive parameters in a layer and therefore, we give importance to these parameters during centroid determination.
The work by \cite{ECSQ} compresses the Mobilenet model by 10.7x using universal vector quantization and universal source coding technique. The \cite{HAQ} paper proposed to use reinforcement learning to automatically determine the quantization policy that's best based on the hardware accelerator's feedback and test their algorithm on MobileNetv1,v2 and ResNet models. In another paper \cite{GOBO}, the authors propose the GOBO algorithm for weight clustering using a method similar to k-means on NLP models. In \cite{Permute-quant-finetune}, the authors propose to permute the weights before compressing and fine-tuning the layers and this results in significant performance improvement. The authors in \cite{cv-sensitivity-comp} propose a non-retraining setup for compression of vision models by determining the best k based on accuracy loss. Whereas, in our B\&Q work, we focus on a novel algorithm for compression based on the sensitivity of the network to noise and the method is applicable to both float and integer quantized models.

Finally, in order to deploy the models in INT DSPs, both the weights and activations should be quantized to INT values. 
Model quantization methods have also gained popularity due to their ability to quantize both the weights and activation values \cite{Zhao2020Linear}. In \cite{EWGS}, the authors propose EWGS as an alternative algorithm for a straight-through estimator(STE) by scaling each gradient element based on the sign of the gradient and the discretizer error. They showed an accuracy of 73.9\% on the imagenet dataset using the ResNet34 model with 4-bit weights and 4-bit activations. In \cite{QGT}, again an algorithm was proposed for training quantized models by using a customized regularization loss for directing the weight values towards a distribution with maximum accuracy while minimizing quantization error. On MobilenetV2 for the person detection task, they compressed it to 2 bits and showed 87.5\% accuracy. PROFIT \cite{PROFIT} is another Mobilenet model compression technique where the activation instability due to weight quantization is alleviated by progressive freezing and iterative training of the model. They achieved 4-bit compression of the Mobilenetv1 model for a 60.056\% accuracy. The authors in \cite{Additive-power-two} presented an additive power of two quantization technique. Their reparametrization of the clipping function is applied for a better-defined gradient for learning the clipping threshold and weight normalization for stabilizing the training. They showed ResNet-18 compressed to 5-bit weight and activation with 70.9\% accuracy on the ImageNet dataset. Memory efficient networks such as MobileNet \cite{mobilenet1,mobilenet2} models, EfficientNet \cite{efficientNet, efficientNet2} are alternatives for directly training efficient neural networks instead of training a heavy network and then compressing and/or quantizing them. However, in our study, we compress and quantize the memory-efficient Mobilenet network as well. 

Although all the model quantization papers focused only on weight and activation quantization technique, \cite{Quant-Noise} authors presented a method for both storage compression and weight/activation quantization using product quantization. Their algorithm randomly quantizes different random subset of weights during each forward pass rather than all the weights as a quantization-aware training scheme and improved the performance of the model while inducing extreme compression. But the authors only tested their method on the EfficientNet-B3 model and RoBERTa model and reduced the representation to 4-bit weights and activation for a significant loss in accuracy. 

In the case of speech recognition models, there have been several works on compression. In \cite{47446-comp-end-to-end}, they used knowledge distillation to train the student network. However, the knowledge distillation method is resource intensive. The other works \cite{47446-comp-end-to-end,prabhavalkar2016compression} applied Singular Value Decomposition (SVD) and pruned the neurons for network compression. The authors in \cite{mcgraw2016personalized} showed compression to 1/3rd the original size but they had to use model adaptation and fine-tuning for maintaining the network performance. Recently, in \cite{speech-lottery} trained highly sparse speech recognition networks that were 20\% of the full weight models and the network also exhibited noise robustness. The authors in \cite{zeroshot-speech} proposed to quantize the speech recognition models and calibrate the model during quantization using synthetic data. 

\begin{figure*}[!h]
\centering
\begin{minipage}[c]{0.4\linewidth}
\includegraphics[width=\linewidth]{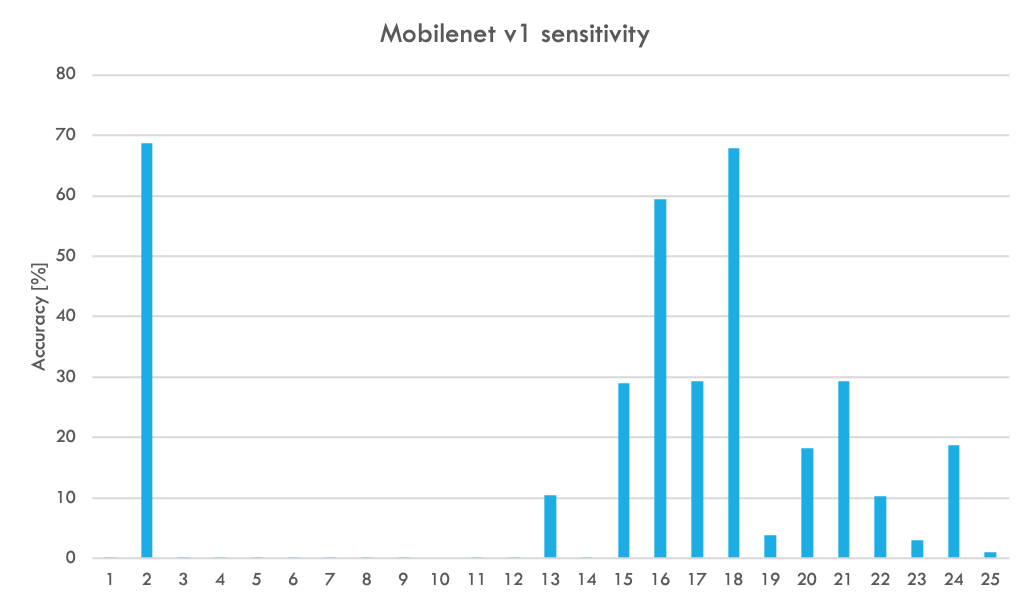}
\end{minipage}\hfill
\begin{minipage}[c]{0.6\linewidth}
\includegraphics[width=\linewidth]{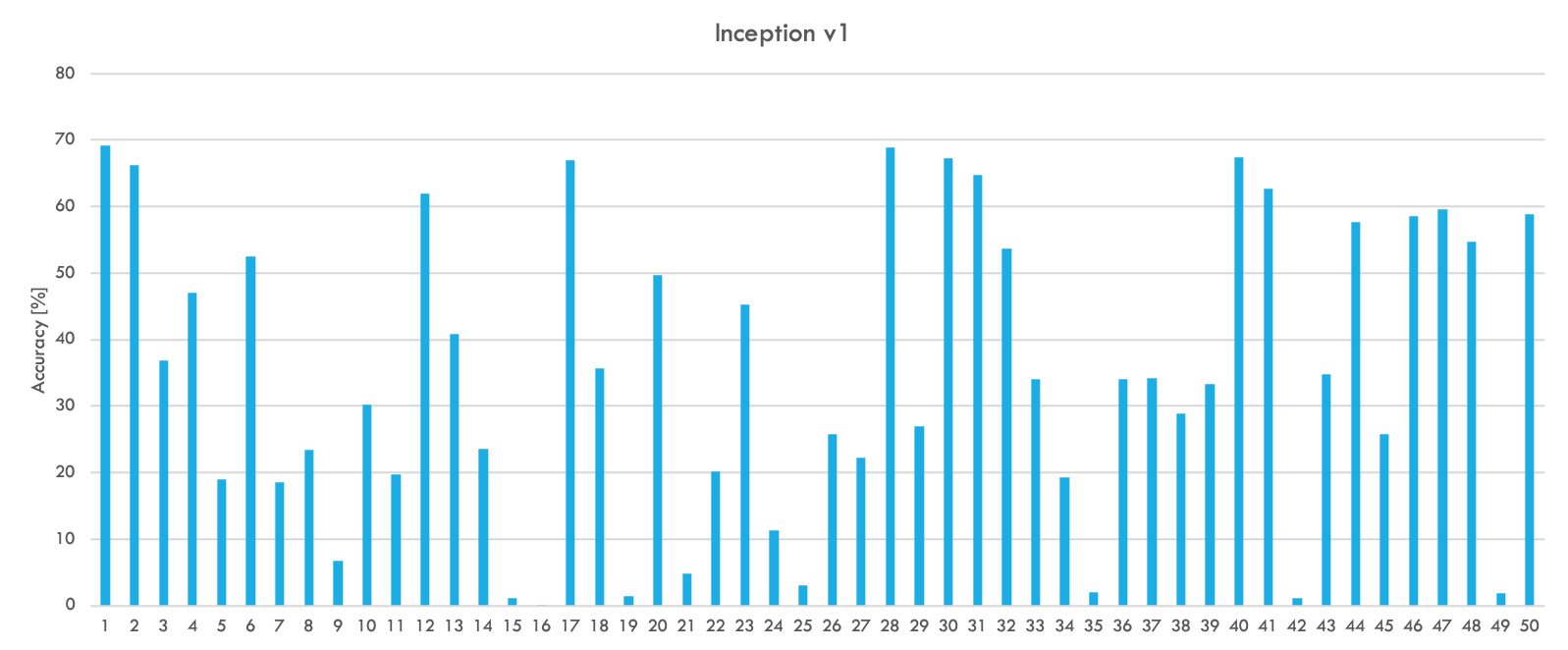}
\end{minipage}\hfill
\\
\begin{minipage}[c]{0.4\linewidth}
\includegraphics[width=\linewidth]{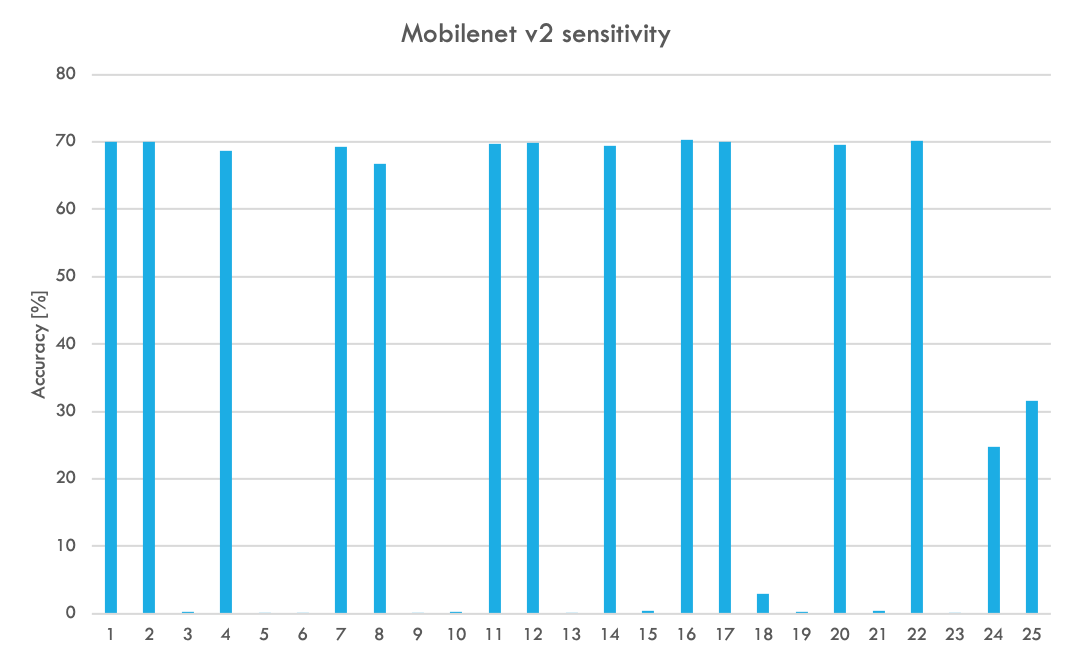}
\end{minipage}\hfill
\begin{minipage}[c]{0.6\linewidth}
\includegraphics[width=\linewidth]{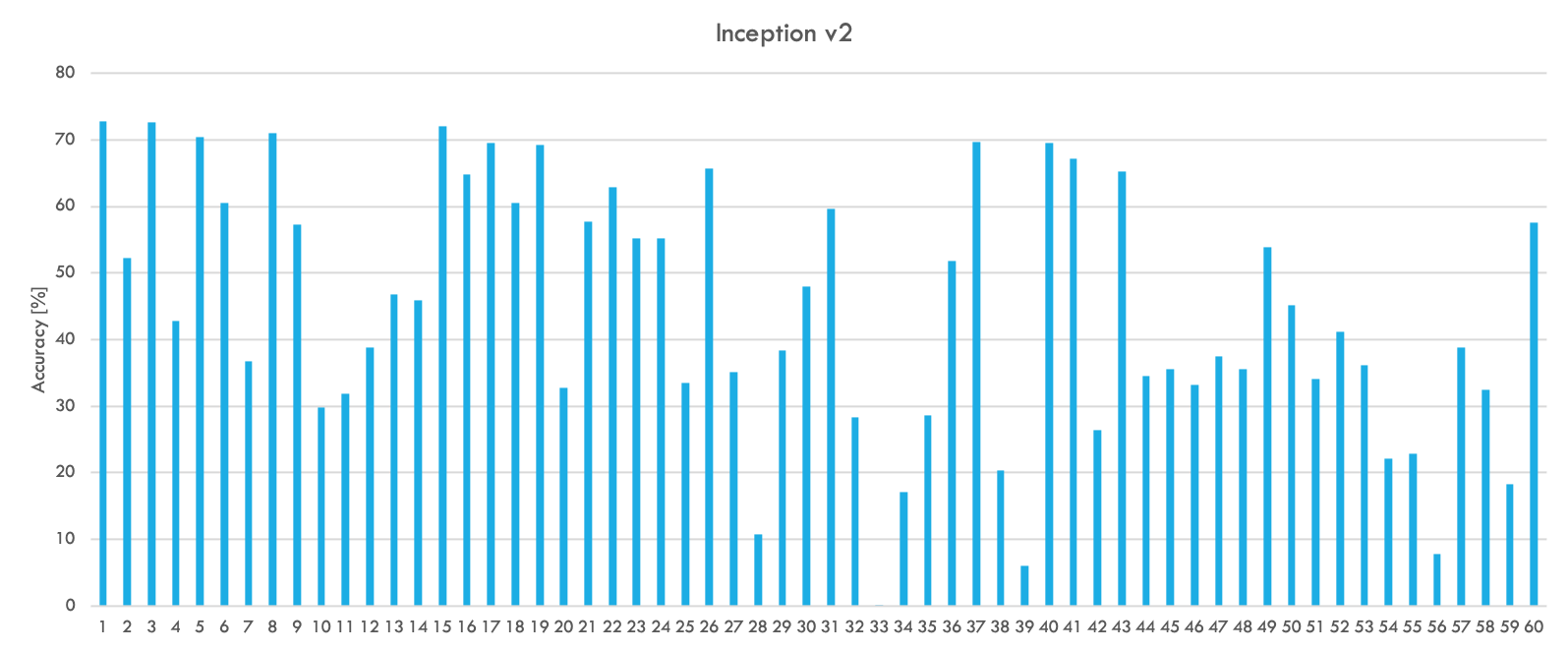}
\end{minipage}\hfill

\caption{From left to right, the effect of magnitude perturbation is shown across 4 quantized models trained on the ImageNet dataset.}
\label{int8Perturb}
\end{figure*}

\section{Method}
\subsection{Bin \& Quant}
\begin{figure}[h]
\begin{center}
\includegraphics[width=0.6\linewidth]{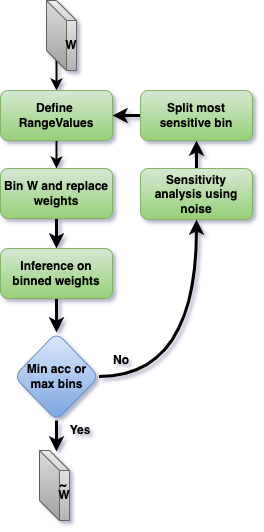}
\centering
\caption{Overall Bin \& Quant algorithm based on sensitivity analysis.}
\label{Algo-BQ-img}
\end{center}
\end{figure}
We designed the Bin \& Quant method \cite{icassp} specifically for compressing speech recognition models. We compress Speech Command and Control (SPC), Deep Speech, Deep Speech 2 and VGG16 models using the proposed algorithm. 
In this scheme, we used a multi-stage pipeline for compression. First, we identify the parameter-intensive layer in each network. Next, we performed a sensitivity analysis on each layer to identify the layers that we could compress together. The sensitivity of a layer is measured as the increase in error rate as a function of perturbation. In DS 2\cite{amodei2015deep} and VGG16\cite{simonyan2014deep}, we compressed more than one layer together. We observed that even after several epochs of training, the trained model weights, when plotted on a histogram, roughly assume a Gaussian distribution. Therefore, we split the weights into 8 bins using the inverse of the standard normal cumulative distribution function. This generated bins, denser around the center and sparser around the ends of the distribution. To each of these bins, we added Gaussian noise(GN) with standard deviation(STD) derived through the sensitivity analysis. The highly sensitive bins were again split in the middle and checked for sensitivity. We continued this process either until the desired number of bins was achieved or if the performance did not drop below 2\% for each bin value. Finally, we used a bit array storage technique for storing the indexes. The index of each parameter were represented using 5 bits since there were at max 20 unique bin values. Using our method, we achieve a storage reduction of $32nm/(log(b)nm + 32b)$, where b is the total number of bins, $n$ is the size of the row and $m$ is the size of the column.
In \cite{DC}, they performed k-means clustering on each layer to find individual cluster values for each layer. The k parameter has to be tuned as a hyperparameter. In \cite{zhou2017balanced}, they performed histogram equalization again on each layer and identified the bin values under the condition that each bin should have equal number of parameters. Whereas, we performed sensitivity analysis and identified the number of bins based on the drop in performance without having the condition that each bin should have equal parameters.


The motivation behind performing sensitivity analysis is to identify the most important neurons in each network. This would help in compressing the other neurons in the network with a higher compression rate. But, with millions of parameters, it is computationally infeasible to identify the important neurons by perturbing each neuron and measuring its effect on the accuracy. In the first set of sensitivity analysis, we add Gaussian Noise on the network layer that contributed to most of the storage to understand its sensitivity. In the second set of analyses, we binned layers into 8 coarse bins. The bin values were obtained using the inverse cdf function. We added magnitude-based perturbation, $\pm 50\%$ of the original weight value and performed an inference cycle to find the accuracy of the perturbed model. The parameters were considered sensitive if there is a drop in accuracy.  

In figure \ref{GaussBin-layerPerturb}, the first row shows the effect on model performance due to Gaussian noise. We perturbed the layers with most parameters and the layers around it. The VGG16 and SPC is measured using Accuracy, DS\cite{hannun2014deep} and DS 2 is measured by the error rate. In each figure, from left to right, we have the layer progressing towards the final output. Each network is perturbed with different STD based on their error tolerance and the range of values in each weight matrix. In the case of VGG16, we noticed that the middle FC layer was least perturbed by Gaussian noise and the FC layer overall is less sensitive than the CNN layer. We noticed similar behavior in the case of the SPC\cite{43969-speechCommand} model. The CNN layers were more sensitive than the FC layer. In the DS model we noticed that for Gaussian noise less than 0.1 STD, the Bi-LSTM layer was more sensitive than the FC layers. Also, the FC layers at the start of the network were more sensitive than the layers towards the end. 
In DS 2, we perturbed the CNN, RNN and FC layers. The RNN layer overall was more sensitive than the FC and CNN layers. Also, the RNN layers towards the end of the network were more sensitive than the layers before that.

The figure \ref{GaussBin-layerPerturb} the second row shows the sensitivity of each bin in each layer for a $\pm 50\%$ relative magnitude perturbation. The parameter distribution was split into 8 bins for VGG16, DS ,and DS 2 models and into 4 bins for SPC model. Since the trained model parameters roughly assume a Gaussian distribution, we used the inverse cdf function to find the 8 bin values for a Gaussian distribution. We perturbed each bin in each layer with $\pm 50\%$ relative magnitude based on the original weight. 
Also, the common notion is that the bins on the end of the distribution would be less sensitive than the middle because the ends would have relatively fewer parameters than at the center of the distribution. In VGG16, middle FC layer and SPC, CNN layer the bins on the ends were more sensitive than the bins at the center. However, that is not the case for the DS model. Therefore, having a finer bin not just at the center but also at the end of the distribution is essential to retain the performance during compression for certain models. 
The CNN layer is more sensitive than the FC layer in SPC model. In DS 2, the 4th bin's sensitivity decreased as we go down the RNN layer. The initial layers seem to be more prone to perturbation compared to the layers after them and finally, the middle RNN layer is the least perturbed of all. Across all the models, the bins at the initial layers were more sensitive than the layers towards the output. Therefore, from sensitivity analysis, we noticed that the type of neuron and the location in the network play an important role in determining its sensitivity. Typically, the final layer is not always the most sensitive. The ends of the distribution are sometimes more sensitive than the middle of the distribution and would require finer bins to retain the model accuracy after compression.

\subsection{Bin \& Quant for Quantized models}
Quantized models are generally derived from their parent original floating point models by applying post-training quantization to weights and/or activations. This technique helps in reducing the model size while improving CPU and hardware accelerator latency. Therefore, this is a preferred method for compressing the model size. Traditionally, in post-training quantization, static quantization is applied where the activations and weights are statically quantized using a calibration dataset. However, this method could lead to a significant drop in accuracy. To combat this issue and yet achieve quantization, the quantization-aware training technique is used, in which the activations and weights are fake quantized and finetuned using the dataset to achieve good accuracy even with weights and activations being quantized.



This proposed algorithm for compression of the quantized model is similar to the original Bin \& Quant algorithm which was applied to the float models. Since the models were already quantized, there is an inherent 4x compression compared to the float model. Here, we apply the storage compression method directly to the quantized weight values which retain the int computation advantage and provide an extra reduction in model size without retraining all the model parameters. Therefore, this compression scheme is computationally less expensive than other schemes that involve retraining the model. Once the layer with the most number of parameters is identified in descending order, we sequentially apply the B\&Q to each layer and observe the drop in accuracy vs. compression gains. We set a particular drop in accuracy that is expected after compressing a given layer and if the accuracy drop exceeds the threshold, that layer will not be compressed and the following layers will be compressed based on their accuracy drop. This way, we avoid manually identifying the most sensitive layer and avoiding compression but rather automate the process based on a desired final accuracy. In each layer, the initial bins are determined based on the mean and std of the parameters. Every bin will then be perturbed by random noise and the bin with the least accuracy after the perturbation is the most sensitive hence that bin is split at the center into two new bins, thereby gaining a finer representation of the values in the bin. This process continues until the maximum allowed bins are reached for each layer and the given number of layers has been tested for compression. 

We again applied sensitivity analysis to mobilenetV1, mobilenetV2, InceptionV1 and InceptionV2 models. We perturbed each layer with a $\pm 0.03$ times the original value as noise. Compared to all the models, mobilenetv1 was the most sensitive. In figure \ref{int8Perturb} we show the accuracy of the models after perturbing each layer one at a time and it is listed as layers with the most number of parameters to the least number of parameters. Except for the mobilenetV1 model, across all the other models, the layer with the most number of parameters was the least sensitive. Also, the next layer in this order was also less sensitive than the following layers except for the inceptionV2 model. This particular empirical result emphasizes the need to understand the capability of compressing each layer irrespective of the number of parameters. Generally, it cannot be concluded that the layer with the most number of parameters is highly redundant and could be compressed easily. The sensitivity of the mobilenetv1 layers also explains the least compressibility of the model and the drop in accuracy due to compression. Since mobilenet models are already designed as an efficient network using point-wise convolution, between mobilenet and inception networks, the sensitivity analysis highlights that mobilenet is more prone to accuracy drop in the presence of noise than inception networks. Also, between mobilenetv1 and mobilenetv2 models, the mobilenetv1 model has additional layer perturbations resulting in almost zero accuracy compared to the mobilenetv2 model and therefore, mobilenetv2 is more compressible than the mobilenetv1 model.

\subsection{Significance of gradients for compression}
As the compression complexity increases, the ability to obtain the utmost compression for the least loss in accuracy becomes more viable. Therefore, in cases where retraining of the model is possible for the purpose of compression, we present an algorithm called Gradient Weighted K-means (GWK). However, the motivation behind using gradients to guide the compression stems from the concept that, during backpropagation for a given batch that is being trained, there would be a higher gradient i.e., a larger change to be applied to a particular weight value based on how well that weight value aided in predicting a particular class for a given input. Therefore, we hypothesized that, after completely training the float model, during an extra training epoch phase, the weight values with a corresponding higher gradient tends to be more sensitive. That is, the trained value should be preserved to the maximum possible extent in order to avoid a degradation in performance while the model is being compressed. 

\begin{figure*}[h]
\centering
\begin{minipage}[c]{0.33\linewidth}
\includegraphics[width=\linewidth]{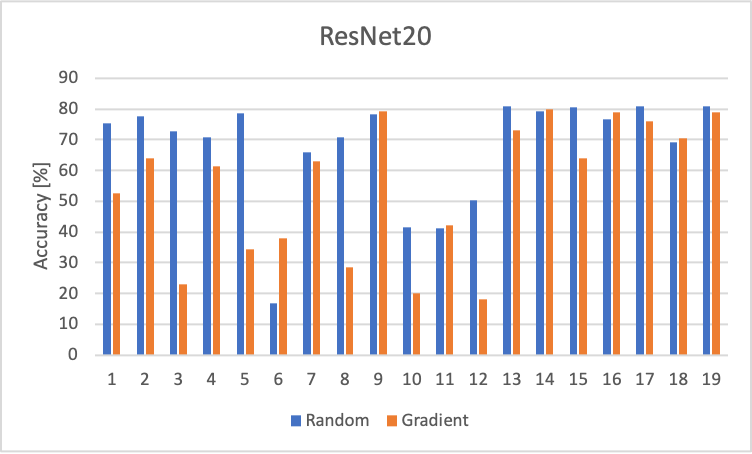}
\end{minipage}\hfill
\begin{minipage}[c]{0.33\linewidth}
\includegraphics[width=\linewidth]{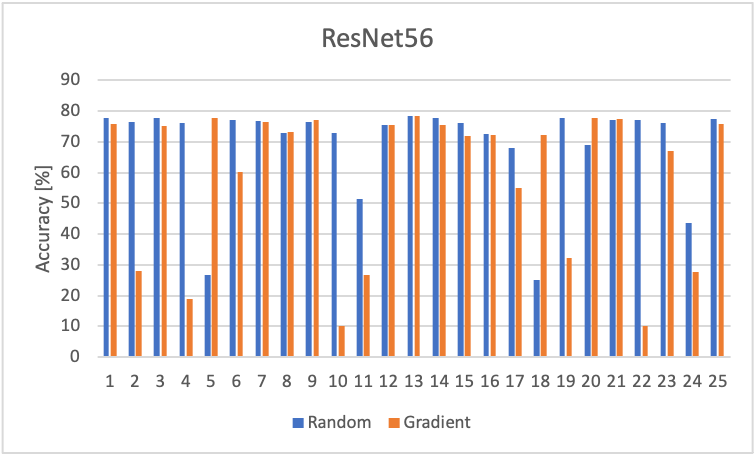}
\end{minipage}\hfill
\begin{minipage}[c]{0.33\linewidth}
\includegraphics[width=\linewidth]{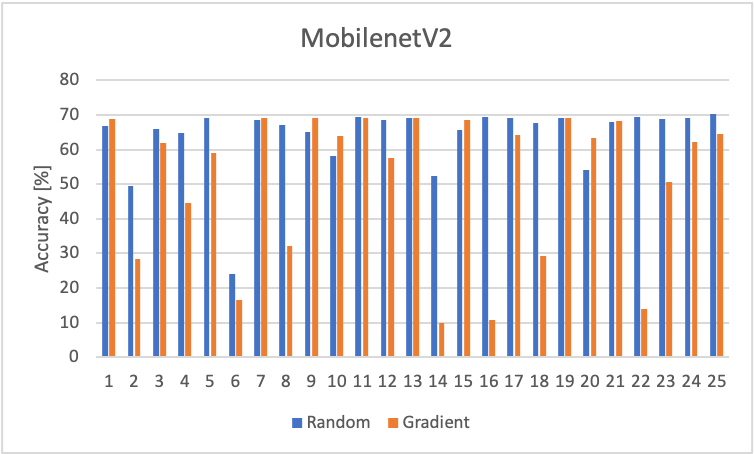}
\end{minipage}\hfill

\caption{From left to right, the effect of random Gaussian noise is shown across 3 models, ResNet20, ResNet56 \& MobileNetV2 trained on the CIFAR-10 dataset.}
\label{GWKPerturb}
\end{figure*}

In order to empirically check this hypothesis, we present a gradient-based sensitivity analysis technique. In this method, the trained model is perturbed one layer at a time with random Gaussian noise of 0 mean and 0.05 std. At each iteration of perturbing a single layer, in case of random perturbation, 50\% of the weights in that layer will be randomly chosen and Gaussian noise would be added. In the case of gradient perturbation, the top 50\% of the weights with the most absolute value in gradient will be added with the same Gaussian noise (0 mean and 0.05 std). In figure \ref{GWKPerturb}, we show the change in the accuracy of the model after applying noise either randomly or using gradient information. 
All three models, ResNet20, ResNet56 and MobileNetV2 are trained and tested on the CIFAR-10 dataset. Similar to quantized model sensitivity analysis, the weight values are arranged in descending order based on the number of parameters. For each layer, in a separate iteration, noise is added randomly and based on gradients and the inference accuracy on the validation set is measured. 
In the case of ResNet20, in the first top parameters layer, the drop in accuracy based on gradient perturbation is higher than that of random perturbation. This trend is followed by all the convolution layers except for 4 layers out of the total 19 other layers. Therefore, in this model, the gradient information is a good measure of the sensitivity of the parameters in a given layer. The ResNet56 model was also similarly perturbed using the Gaussian noise and the top 25 layers are plotted in figure \ref{GWKPerturb}. Layers 2,4,10,11 and 22 show a significant drop in accuracy when perturbed using the gradient information vs. when perturbed randomly. This again highlights the use of the gradients for identifying the most sensitive parameter. Finally, in the case of the mobilenetV2 model, layers 2,8,14,16,22 also show significant degradation in accuracy. In this model, almost all layers with most parameters were point-wise conv layers. Therefore, the gradient-based sensitivity method is applicable on both point-wise and regular convolution layers.  

\subsection{Gradient weighted k-means - GWK}
\begin{figure}[h]
\begin{center}
\includegraphics[width=\linewidth]{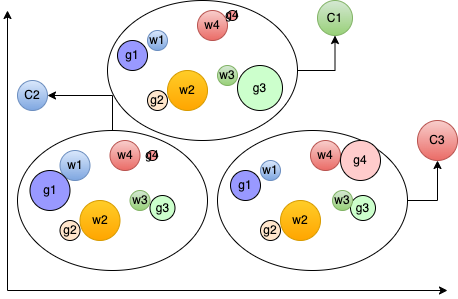}
\centering
\caption{\label{clustering}Gradient-weighted k-means clustering of weights based on gradient value.}
\end{center}
\end{figure}
Given the significance of gradients in identifying the most sensitive parameter, we propose an algorithm to guide cluster formation based on the gradient weights. In previous literature, the focus has either been to compress the weight value or quantize the weight value. The combination of both techniques \cite{Quant-Noise} has only been tested until 4-bit activation. In our algorithm, we compress the weight values using product quantization and gradient-weighted k-means clustering and retrain the network using EWGS \cite{EWGS} to train and compress the quantized network. To the best of our knowledge, we are the first ones to combine storage compression using gradients and network quantization technique using the latest EWGS algorithm for both effective inference speed-up (Quantization) and reduction in the size of the model. 
Overall, the algorithm starts with the pre-trained floating point model and the network will be trained using EWGS algorithm. 

After each epoch, the weight values will be rearranged into a 2-d matrix based on block size for product quantization and gradient-weighted k-means will be applied on the rearranged 2-d matrix.
Similar to \cite{ATB}, a given convolution weight matrix $\textbf{W} \in \textbf{R}^{C_{in} \times C_{out} \times k \times k}$ is reshaped to a 2-D matrix $\textbf{W} \in \textbf{R}^{(C_{in}  \times k \times k) \times C_{out}}$ and then rearranged to the final 2-D matrix of size $\textbf{W} \in \textbf{R}^{d \times n_{blocks}}$. The block\_size is the variable $d$. This is a hyperparameter represented as cv in the results table and the pw refers to the $d$ allocated for point-wise convolution. The grad values will now be rearranged as a 2-d vector similar to weights, $\textbf{g} \in \textbf{R}^{d \times n_{blocks}}$. However, for k-means, the first dimension is the sample and the second dimension is the feature space. So, across the first dimension, the absolute value of the gradient will be averaged and a final normalization is applied to the 1-d gradient vector. These two vectors, weights and gradients are given to k-means clustering with a set n\_cluster value. 
Every candidate centroid $c_j$ for the weight $W$ is derived using,
\begin{equation}
    c_j = \frac
    {\sum_i g_i w_i} 
    {\sum_i g_i}
\end{equation}
This generates centroids of size $\textbf{C} \in \textbf{R}^{d \times b}$, where $b$ is the total number of clusters and labels of size $\textbf{L} \in \textbf{R}^{n_{blocks}}$.
The output of the k-means algorithm, centroids and labels will be saved and decompressed during inference. The weight, $\stackrel{\sim}{\textbf{W}}$  reconstructed using centroids and labels is then used for continuing the training process. At the last epoch, the final weight values will be determined based on the latest gradients and the centroids are fixed for storage compression. 

\begin{table*}[t]
  \centering
  \begin{tabular}{|l|l|l|l|l|l|}
  \hline
    Model & Compression scheme & layer compressed & compressed size & Performance [\%] & Compression\\
    \hline
    \hline
    \multirow{3}{4em}{SPC} & original & - & 3.7 MB & Accuracy 87.0 & - \\
    & k-means & FC & 1.74 MB & Accuracy 85.2 & 2.12x \\
    & \textbf{B\&Q} &FF, Conv & 270.7 KB & Accuracy 85.4 & \textbf{13.66x} \\
    \hline
    \multirow{3}{4em}{DS 1} & original & - & 181 MB & WER: 18.62 CER: 9.23 &  - \\
     & k-means & Bi-LSTM & 169 MB & WER:20.93 CER: 10.73 &  1.11x\\
     & \textbf{B\&Q} & Bi-LSTM & 73 MB & WER: 19.99 CER:10.01 & \textbf{2.47x} \\
    \hline
    \multirow{3}{4em}{DS 2} & original & - & 158 MB & WER: 11.481 CER: 3.075 & - \\
    & k-means & RNN-0,1 & 108 MB & WER: 23.631 CER: 7.588 & 1.5x \\
    & \textbf{B\&Q} & RNN -0,4 &  22 MB & WER: 13.109 CER: 3.499 & \textbf{7.18x}\\
    \hline
    \multirow{3}{4em}{VGG16} & original & -  & 530 MB & Acc@1 71.59 Acc@5 90.38 & - \\
     & k-means & FC-0  & 233 MB & Acc@1 63.476 Acc@5 85.546& 2.27x \\
    & \textbf{B\&Q} & FC-0,2 & 116 MB & Acc@1 71.048 Acc@5 90.228 & \textbf{4.56x} \\
    \hline
  \end{tabular}
  \caption{The results of three compression techniques applied on speech recognition and vision models are shown in the table.}
  \label{tab:1}
\end{table*}

During the training process, the latest EWGS \cite{EWGS} algorithm shifts each gradient value based on a scaling factor derived using the Hessian information of the network. The authors have shown that for quantized weight training, EWGS is more effective compared to the STE algorithm. Therefore, in our algorithm, since the goal is to induce both storage compression and quantization, we use the EWGS for training our clustered/compressed weight values to achieve the utmost compression.  
In figure \ref{clustering}, we show an illustration of weight clustering based on the gradient values. Given three cluster regions, in the case of cluster 1, the gradient of weight w3 has the highest magnitude and despite the value of w2 is high, since the gradient of w2, g2 is lower in magnitude, the cluster centroid will be more towards the w3 value rather than w2. Similarly, for the second cluster, the weight is skewed towards w1 and in the case of the third cluster, it is skewed towards w4.

\section{Experiment}
\subsection{Compression using B\&Q}
We used four Deep learning models for our B\& Q experiments. They vary in size, application and also in the type of neurons utilized to construct the model. 
The SPC model is a 10-word speech recognition system. This network was trained using the Speech Command and Control Dataset\cite{warden2018speech}. The original model size of this network is 3.7 MB with 2 CNN layers and 1 FC layer. The accuracy of this model is 87\%. Deep Speech(DS) is a continuous speech recognition system trained on Fisher\cite{Cieri2004TheFC-fisherDataset}, LibriSpeech\cite{Panayotov2015LibrispeechAA}, Switchboard\cite{225858-switchboard} pre-release snapshot of the English Common Voice training corpus. The original model size is 181 MB with a Word Error Rate(WER) of 18.62\% and Character Error Rate(CER) of 9.23\% on the LibriSpeech clean test set. The network consists of 5 FC hidden layers and a Bi-LSTM layer. Deep Speech 2 (DS 2) is a continuous speech recognition system trained on the LibriSpeech (1000 hours) dataset. It consists of CNN, Bi-GRU and a FC layer. The original model size is 158MB and it yields a WER of 11.481\% and CER of 3.075\% on the LibriSpeech test clean dataset. VGG16 is an object classification network trained on the ImageNet dataset with 1000 classes of images. It consists of 13 CNN layers and 3 FC layers. The original model size is 530 MB with Top-1 accuracy of 71.59\% and Top-5 accuracy of 90.38\% on the Imagenet validation dataset. 
The compression result using B\&Q on floating point models are reported in \ref{tab:1}. K-means clustering and B\&Q are applied to all the models. The k-means cluster size is determined based on the best compression vs. accuracy trade-off. In the case of the SPC model, we set 512 as the k-means cluster size and for the B\&Q method, the initial bins were found using the inverse CDF function with 0 mean and 0.025 std. Based on sensitivity analysis we found 4 bins to be ideal for the desired accuracy and only 2 bits were required for storage of the index of each weight value. This resulted in a 13.66x compression ratio for a final 85.4\% accuracy while the original model accuracy was 87.0\%. Although the final accuracy of the k-means method was also 85.2\%, this method resulted only in a 2.12x compression ratio. 
The Deep Speech model was clustered with a k-means cluster size of 3500 and the initial B\&Q bins were derived based on 0 mean and 0.5 std. The B\&Q algorithm with sensitivity analysis resulted in 20 bins and hence 5 bits were used to represent the index value. In this case, only the Bi-LSTM layer was compressed for a final 2.47x compression ratio and 10.01\% CER. The k-means algorithm resulted in 10.73\% CER, similar to B\&Q but was only compressed to 1.11x relative to the original model size. Among the models that were compressed using B\&Q, this was compressed the least in order to maintain the least loss in accuracy. 

Deep Speech 2 model compressed using k-means clustering used 512 clusters. The B\&Q with initial bins based on mean 0 and std 0.5 resulted in 16 bins and hence 4 bits were required for storing. Additionally, since the bin values across all RNN layers had similar bin values, we established parameter sharing across layers. The resulting WER was 13.109\% and CER was 3.499\% with a 7.18x compression and 22MB compressed size. In another case, the RNN-2 layer with only 3 bits and the rest of the layers with 4 bits resulted in 13.759\% WER and 3.675\% CER for a final size of 21 MB, 7.52x compression. Therefore, it is a constant trade-off between the size of the model vs. its accuracy. While the k-means algorithm resulted in a model with 7.588\% CER significantly higher than B\&Q for only a 1.5x compression ratio. Although k-means has been successfully applied for weight clustering \cite{DC}, they always require re-training to regain the lost accuracy due to clustering. But, in our algorithm, we propose to not retrain the models in case of resource constraints and yet B\&Q is an effective way to compress the models without retraining. 
In the case of VV16, k-means with 1000 clusters resulted in 63.476\% top-1 accuracy and B\&Q with mean 0 and 0.01 std resulted in 12 bin values, 4 bits per compressed weight value. The final compressed model resulted in 71.048\% top-1 accuracy at a 4.56x compression ratio. 
The std for inverse CDF across all models was derived based on the range value of the parameters in a layer. 
Across all models, our B\&Q method gave the best trade-off between accuracy and compression.

\begin{table*}[t]
  \centering
  \begin{tabular}{|l|l|l|l|l|l|l|}
  \hline
    Model & Compression scheme & compressed size &  Performance [\%] & W/A & Retraining & Compression \\
    \hline
    \hline
    \multirow{3}{5em}{MobileNetV1} & original & 4.07 MB & 68.11 & 8/8 & - & -\\
     & DKM \cite{DKM} & 0.72MB & 63.9 & 32/32 & Yes & 5.59x\\
     & k-means & 3.39MB  & 63.28 & 8/8 & No & 1.19x\\
     & \textbf{B\&Q} & 3.08MB & 64.42 & 8/8 & No & 1.38x\\
    \hline
    \multirow{4}{4em}{MobileNetV2} & original  & 3.5 MB & 69.64 &  8/8 & -& -\\
     & WS \cite{WS}  & 3.18MB & 71.9 & 32/32 & No & 1.1x \\
     & DKM \cite{DKM}  & 0.84MB & 68.0 & 32/32 & Yes & 3.95x \\
     & k-means & 2.25MB  & 46.08 & 8/8 & No & 1.53x\\
     & \textbf{B\&Q} & 2.24MB & 66.68 & 8/8 & No & 1.58x \\
    \hline
    \multirow{2}{4em}{InceptionV1} & original  & 6.37MB & 68.18 & 8/8 & -& -\\
     & k-means &  4.81MB & 63.75 & 8/8 & No & 1.38x\\
     & \textbf{B\&Q} &  4.81MB & 64.16 & 8/8 & No & 1.41x\\
    \hline
    \multirow{2}{4em}{InceptionV2} & original   & 10.74MB & 72.01 & 8/8 & -& -\\
    & k-means &  9.41MB & 67.5 & 8/8 & No & 1.29x\\
    & \textbf{B\&Q} &  8.89MB & 69.48 & 8/8 & No & 1.3x \\
    \hline
    \multirow{2}{4em}{Person detect} & original  & 230KB & 86.13 & 8/8 & -& -\\
    & PTQ \cite{QGT} & 123 KB & 72.2 & 4/8 & No & 2x \\
    & QGT \cite{QGT} & 123 KB & 87.9 & 6/8 & Yes & 2x \\
    & k-means & 151 KB  & 76.36 & 8/8 & No & 1.53x\\
    & \textbf{B\&Q}  & 151KB & 83.89 & 8/8 & No & 1.56x \\
    \hline
  \end{tabular}
  \caption{The results of three compression techniques applied on vision models are shown in the table. This compression ratio is calculated based on quantized models. For comparison with float32 models, the compression factor should be multiplied by 4x.}
  \label{tab:2}
\end{table*}

\begin{figure}[h]
\begin{center}
\includegraphics[width=\linewidth]{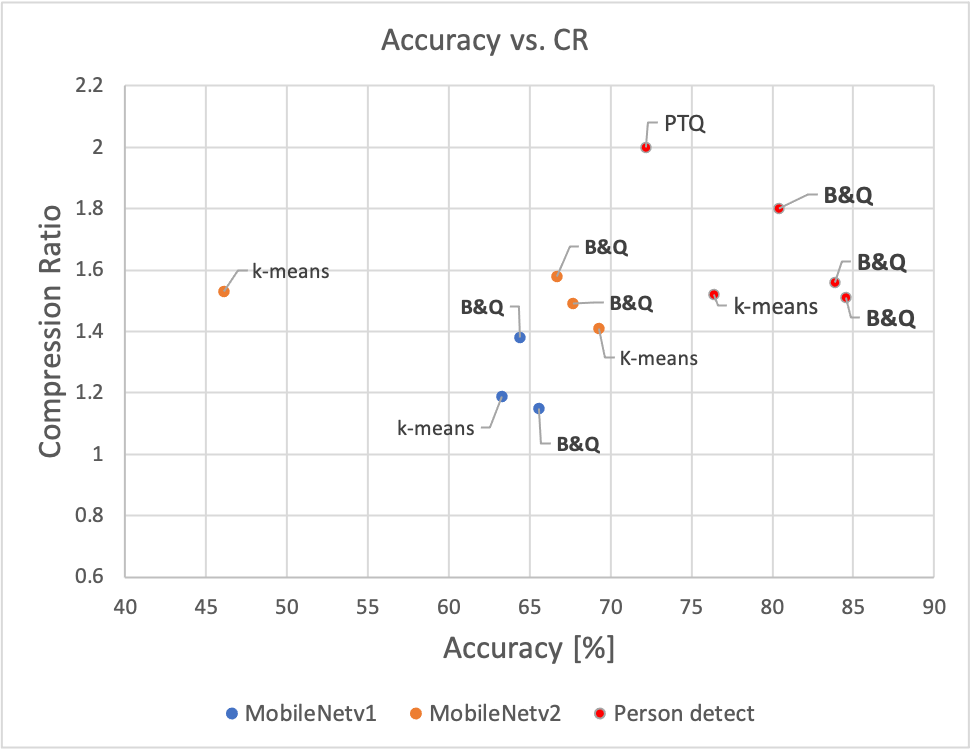}
\centering
\caption{Accuracy vs. Compression Ratio Trade-off. }
\label{Algo-CR_Acc}
\end{center}
\end{figure}

\begin{figure*}[!h]
\centering
\begin{minipage}[c]{0.5\linewidth}
\includegraphics[width=\linewidth]{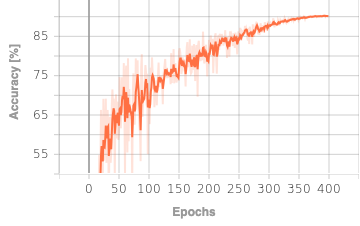}
\end{minipage}\hfill
\begin{minipage}[c]{0.5\linewidth}
\includegraphics[width=\linewidth]{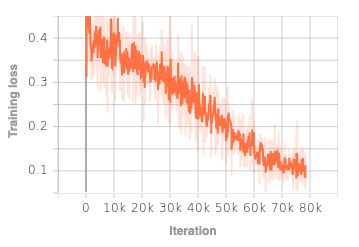}
\end{minipage}\hfill
\caption{The test accuracy and training loss for ResNet-20 model trained using GWK on CIFAR10}
\label{GWK-acc-loss}
\end{figure*}

\subsection{Compression of Quantized models using B\&Q}
The B\&Q for quantized models were directly tested on the publicly available tensorflow quantized tflite models with both weights and activation quantized 8-bits. All the models were trained and tested on the Imagenet dataset with 1000 classes. The algorithm was automated to identify the best bins for each layer and a layer would not be compressed if there was more than a 1\% drop in accuracy and the following layers will be compressed as a continuous process until the total layers are compressed parameter is reached with a sample size of 1000 for all the models except person detect which was set to 4000. In most models, for the first compressed layer, the initial bins were assigned by manually inspecting the histogram of the parameter values. Finally, Huffman compression is applied to the index table and the final model sizes are reported. The result across various models are shown in table \ref{tab:2}. In the case of the k-means baseline, we set the eval size i.e, the number of samples used for evaluation of the sensitivity of the network to 1000 for all the models except person detect which was set to 4000. All the compressed layers were uniformly allocated k=8, for 3-bit representation and the number of layers compressed for each model is based on the desired compression ratio to compare with B\&Q.  
In the case of the mobilenetv1 model, the original uint8, quantized model of size 4.07 MB is compressed using the proposed algorithm and results in a 3.08MB model with 2 layers compressed and a final accuracy of 64.42\%. In addition to B\&Q, we also applied Huffman encoding to the labels and further compressed to 1.38x from a 1.32x compression ratio. 
The mobilenetv2 model had an original size of 3.4 MB and 70.61\% accuracy. This model was reduced by a 1.58x compression ratio, resulting in a 2.24MB model with 66.91\% accuracy. As shown in figure \ref{int8Perturb}, the layers of the mobilenetv1 model were very sensitive compared to the mobilenetv2 model and hence mobilenetv2 model could be compressed further compared to the mobilenetv1 model. 

While the Inceptionv1 model with an original size of 4.81MB with 68.18\% accuracy could be compressed by 1.41x for a final accuracy of 64.16\% after compressing 6 layers, the inceptionv2 model could only be compressed by 1.3x after applying B\&Q to 9 layers for considerably less loss in accuracy compared to inceptionv1. Although more layers are compressed, since B\&Q is applied to model parameters in descending order based on the number of parameters, more compressed layers would only lead to a lesser gain in compression factor. The reported results are on 10000 images from the imagenet dataset for the Inception networks. Since the inference on tflite model had to be performed on CPU, we reduced the size of the test set images. 
Finally, we compressed one more lightweight model, person detect based on mobilenetv1. The person detect model could be compressed by 1.56x for a 2.24\% accuracy drop compared to the original model at a final model size of 160KB. 
Therefore, B\&Q can compress both the heavy and lightweight models effectively. 

\subsection{Compression using GWK}
\begin{table}[t]
  \centering
  \begin{tabular}{|l|l|l|l|l|l|l|}
  \hline
    Model & Compression & W/A & Top-1  &  b/w & cv/pw/bits \\ 
    \hline
    \hline
    \multirow{6}{6em}{ResNet-20} & original & 32/32 & 91.72 & - & - \\ 
     & k-means & 8/8 & 24.06 & 1 & - \\ 
     & EWGS\cite{EWGS} & 1/8 & 90.9 & 1 & - \\ 
    & \textbf{GWK} & 8/8 & 90.48 & 0.904 & 6/4/5 \\ 
     & EWGS\cite{EWGS} & 1/1 & 85.59 & 1 & - \\ 
     & DSQ\cite{DSQ} & 1/1 & 84.11 & 1 & - \\ 
     & IR-Net\cite{IR-Net} & 1/1 & 85.4 & 1 & - \\ 
     & \textbf{GWK} & 8/1 & 85.96 & 1.06 & 6/4/6 \\ 
     & \textbf{GWK} & 8/1 & 85.53 & 0.904 & 6/4/5 \\ 
    \hline
    \multirow{4}{4em}{ResNet-56} & original & 32/32 & 93.55 & - & - \\ 
     & k-means & 8/8 & 13.99 & 1 & - \\ 
     & \textbf{GWK} & 8/8 & 92.35 & 0.88 & 6/4/5 \\ 
     & EWGS\cite{EWGS} & 1/8 & 93.57 & 1 & - \\ 
    \hline
    \multirow{4}{4em}{MobileNetv2} & Original & 32/32 & 94.75 & - & - \\ 
     & k-means & 1/8 & 10 & 1 & - \\ 
     & EWGS\cite{EWGS} & 1/8 & 91.5 & 1 & - \\ 
     & \textbf{GWK} & 8/8 & 92.15 & 0.91 & 6/8/6 \\ 
    \hline
  \end{tabular}
  \caption{The results of the GWK compression algorithm on the ResNet-20,56 and MobileNetv2 models trained on the CIFAR-10 dataset.}
  \label{tab:3}
\end{table}
We apply our compression algorithm on ResNet-20, and ResNet-56 models trained on CIFAR10. The baseline compression method is k-means with a fixed bit per weight applied to a given layer directly without product quantization but trained using the EWGS algorithm. 
The other baseline method is the EWGS algorithm applied for a particular bit-width. In the case of the ResNet-20 model, the GWK algorithm reduces the bit/weight to 0.904 (sub-1-bit representation) for a final 90.48\% accuracy. This results in an overall 35x compression ratio for a model with quantized activations. The previous literature \cite{DKM}, shows sub-1-bit representation on the vision networks but with 32-bit activations. 
Although EWGS with 1-bit weight and 8-bit activation resulted in 90.9\% accuracy, the EWGS could not be compressed further due to hardware constraints and our method has the advantage of compression to sub-1-bit representation. 
In the case of 1-bit activation, GWK resulted in 85.96\% final accuracy for 1.06 bits/weight representation while EWGS gave 85.59\% final accuracy for 1 bit/weight storage. Figure \ref{GWK-acc-loss} shows the change in training loss over iterations for the ResNet-20 model trained on CIFAR-10. Although after each iteration, the weights are clustered using k-means after applying product quantization, the network exhibits stable learning beyond 300 epochs. Similarly, the training loss also decreases as the network is trained for more iterations. 

The ResNet-56 model's baseline compression algorithm k-means with 8-bit weight and 8-bit activation gave a final accuracy of 13.99\% using 1-bit clustering. Whereas, GWK with 0.88 bits/weight resulted in 92.35\% accuracy, clearly outperforming the k-means technique. In another case, while the bit representation was increased to 1.04 bits/weight, this resulted in 92.8\% accuracy. Again, this shows the trade-off between storage and accuracy. 

Similarly, we compressed the MobileNetv2 model using our proposed GWK algorithm and this resulted in a model with 92.03\% final accuracy for 0.91 bits/weight and 35.16x compression ratio. The baseline k-means algorithm only resulted in 10\% accuracy and EWGS trained with 1-bit weight and 8-bit activations resulted in a model with 91.5\% accuracy at 32x compression ratio. 

In addition to testing on the CIFAR-10 dataset, we also compressed the MobileNetV2 model trained on the ImageNet dataset for 1000 class classification. The original model has a reported top-1 accuracy of 71.9\% while in the case of DKM \cite{DKM} with 32-bit weight and 32-bit activation model the authors compressed to the model to an effective 2.010 bits/weight for an accuracy of 68.0\%. The model compressed using our GWK algorithm resulted in a 63.8\% top-1 accuracy with 8-bit quantized activations and effective 3.44 bits/weight representation. Although methods such as PROFIT \cite{PROFIT} result in 71.56\% accuracy on a 4-bit weight and 4-bit activation model, it is specifically tested only on the MobileNetV2 models it involves an extensive training routine of progressive training and iterative freezing of layers to achieve this compression. Moreover, our method is effective in compressing the ResNet-20 and ResNet-56 models in addition to MobileNetv2 trained on the CIFAR-10 dataset without requiring knowledge distillation or additional parameters in the training routine and it only relies on pre-computed gradient information used for training in order to determine the best centroid value for a given cluster.

\section{Conclusion}
Since deep learning has been increasingly applied across various fields and for edge applications, it is vital to reduce the storage and compute requirements of the networks. However, in cases where retraining is not always feasible to reduce the model size and yet recover the original algorithm, we present our novel Bin \& Quant method for compressing float models without retraining for a negligible loss in accuracy. In addition to that, in cases where compute must also be saved while storage size reduction is accommodated, we extend our Bin \& Quant algorithm to the quantized models and again show compression for a minimal loss in accuracy without retraining. Hence, using the above technique, the compression pipeline can be swiftly applied to already pre-trained models and compute, storage savings can be achieved. Finally, in cases where it is vital to induce utmost compression at the expense of retraining the model and increasing the compression complexity, we present our novel gradient-weighted k-means clustering (GWK) algorithm and show significant size reduction, sub-1-bit representation on quantized models for a negligible loss in accuracy. We also tested our algorithm on speech and vision models. Therefore, our compression algorithms can be applied for various applications under different training/non-training settings. 

\bibliographystyle{main}
\bibliography{main.bib}

\begin{thebibliography}{10}

\bibitem{EWGS}
Junghyup Lee, Dohyung Kim, and Bumsub Ham,
\newblock ``Network quantization with element-wise gradient scaling,''
\newblock in {\em Proceedings of the IEEE/CVF Conference on Computer Vision and
  Pattern Recognition (CVPR)}, June 2021, pp. 6448--6457.

\bibitem{icassp}
Madhumitha Sakthi, Ahmed Tewfik, and Raj Pawate,
\newblock ``Speech recognition model compression,''
\newblock in {\em ICASSP 2020 - 2020 IEEE International Conference on
  Acoustics, Speech and Signal Processing (ICASSP)}, 2020, pp. 7869--7873.

\bibitem{WS}
Etienne Dupuis, David Novo, Ian O'Connor, and Alberto Bosio,
\newblock ``A heuristic exploration of retraining-free weight-sharing for cnn
  compression,''
\newblock in {\em 2022 27th Asia and South Pacific Design Automation Conference
  (ASP-DAC)}, 2022, pp. 134--139.

\bibitem{DKM}
Minsik Cho, Keivan~A. Vahid, Saurabh Adya, and Mohammad Rastegari,
\newblock ``Dkm: Differentiable k-means clustering layer for neural network
  compression,'' 2021.

\bibitem{ATB}
Pierre Stock, Armand Joulin, Rémi Gribonval, Benjamin Graham, and Hervé
  Jégou,
\newblock ``And the bit goes down: Revisiting the quantization of neural
  networks,'' 2019.

\bibitem{DC}
Song Han, Huizi Mao, and William~J. Dally,
\newblock ``Deep compression: Compressing deep neural networks with pruning,
  trained quantization and huffman coding,'' 2015.

\bibitem{ECSQ}
Yoojin Choi, Mostafa El-Khamy, and Jungwon Lee,
\newblock ``Universal deep neural network compression,''
\newblock {\em IEEE Journal of Selected Topics in Signal Processing}, vol. 14,
  no. 4, pp. 715--726, 2020.

\bibitem{HAQ}
Kuan Wang, Zhijian Liu, Yujun Lin, Ji~Lin, and Song Han,
\newblock ``Haq: Hardware-aware automated quantization with mixed precision,''
\newblock in {\em Proceedings of the IEEE/CVF Conference on Computer Vision and
  Pattern Recognition (CVPR)}, June 2019.

\bibitem{GOBO}
Ali~Hadi Zadeh, Isak Edo, Omar~Mohamed Awad, and Andreas Moshovos,
\newblock ``Gobo: Quantizing attention-based nlp models for low latency and
  energy efficient inference,''
\newblock in {\em 2020 53rd Annual IEEE/ACM International Symposium on
  Microarchitecture (MICRO)}, 2020, pp. 811--824.

\bibitem{Permute-quant-finetune}
Julieta Martinez, Jashan Shewakramani, Ting~Wei Liu, Ioan~Andrei Barsan,
  Wenyuan Zeng, and Raquel Urtasun,
\newblock ``Permute, quantize, and fine-tune: Efficient compression of neural
  networks,''
\newblock in {\em Proceedings of the IEEE/CVF Conference on Computer Vision and
  Pattern Recognition (CVPR)}, June 2021, pp. 15699--15708.

\bibitem{cv-sensitivity-comp}
Etienne Dupuis, David Novo, Ian O’Connor, and Alberto Bosio,
\newblock ``Sensitivity analysis and compression opportunities in dnns using
  weight sharing,''
\newblock in {\em 2020 23rd International Symposium on Design and Diagnostics
  of Electronic Circuits and Systems (DDECS)}, 2020, pp. 1--6.

\bibitem{Zhao2020Linear}
Xiandong Zhao, Ying Wang, Xuyi Cai, Cheng Liu, and Lei Zhang,
\newblock ``Linear symmetric quantization of neural networks for low-precision
  integer hardware,''
\newblock in {\em International Conference on Learning Representations}, 2020.

\bibitem{QGT}
Sedigh Ghamari, Koray Ozcan, Thu Dinh, Andrey Melnikov, Juan Carvajal, Jan
  Ernst, and Sek Chai,
\newblock ``Quantization-guided training for compact tinyml models,'' 2021.

\bibitem{PROFIT}
Eunhyeok Park and Sungjoo Yoo,
\newblock ``Profit: A novel training method for sub-4-bit mobilenet models,''
\newblock in {\em Computer Vision – ECCV 2020: 16th European Conference,
  Glasgow, UK, August 23–28, 2020, Proceedings, Part VI}, Berlin, Heidelberg,
  2020, p. 430–446, Springer-Verlag.

\bibitem{Additive-power-two}
Yuhang Li, Xin Dong, and Wei Wang,
\newblock ``Additive powers-of-two quantization: An efficient non-uniform
  discretization for neural networks,'' 2019.

\bibitem{mobilenet1}
Andrew~G. Howard, Menglong Zhu, Bo~Chen, Dmitry Kalenichenko, Weijun Wang,
  Tobias Weyand, Marco Andreetto, and Hartwig Adam,
\newblock ``Mobilenets: Efficient convolutional neural networks for mobile
  vision applications,'' 2017.

\bibitem{mobilenet2}
Mark Sandler, Andrew Howard, Menglong Zhu, Andrey Zhmoginov, and Liang-Chieh
  Chen,
\newblock ``Mobilenetv2: Inverted residuals and linear bottlenecks,''
\newblock in {\em Proceedings of the IEEE Conference on Computer Vision and
  Pattern Recognition (CVPR)}, June 2018.

\bibitem{efficientNet}
Mingxing Tan and Quoc~V. Le,
\newblock ``Efficientnet: Rethinking model scaling for convolutional neural
  networks,''
\newblock 2019.

\bibitem{efficientNet2}
Mingxing Tan and Quoc~V. Le,
\newblock ``Efficientnetv2: Smaller models and faster training,''
\newblock 2021.

\bibitem{Quant-Noise}
Angela Fan, Pierre Stock, Benjamin Graham, Edouard Grave, Remi Gribonval, Herve
  Jegou, and Armand Joulin,
\newblock ``Training with quantization noise for extreme model compression,''
  2020.

\bibitem{47446-comp-end-to-end}
Ruoming Pang, Tara Sainath, Rohit Prabhavalkar, Suyog Gupta, Yonghui Wu,
  Shuyuan Zhang, and Chung-Cheng Chiu,
\newblock ``Compression of end-to-end models,''
\newblock 2018.

\bibitem{prabhavalkar2016compression}
Rohit Prabhavalkar, Ouais Alsharif, Antoine Bruguier, and Ian McGraw,
\newblock ``On the compression of recurrent neural networks with an application
  to lvcsr acoustic modeling for embedded speech recognition,'' 2016.

\bibitem{mcgraw2016personalized}
Ian McGraw, Rohit Prabhavalkar, Raziel Alvarez, Montse~Gonzalez Arenas,
  Kanishka Rao, David Rybach, Ouais Alsharif, Hasim Sak, Alexander Gruenstein,
  Francoise Beaufays, and Carolina Parada,
\newblock ``Personalized speech recognition on mobile devices,'' 2016.

\bibitem{speech-lottery}
Shaojin Ding, Tianlong Chen, and Zhangyang Wang,
\newblock ``Audio lottery: Speech recognition made ultra-lightweight,
  noise-robust, and transferable,''
\newblock in {\em International Conference on Learning Representations}, 2022.

\bibitem{zeroshot-speech}
Sehoon Kim, Amir Gholami, Zhewei Yao, Nicholas Lee, Patrick Wang, Aniruddha
  Nrusimha, Bohan Zhai, Tianren Gao, Michael~W. Mahoney, and Kurt Keutzer,
\newblock ``Integer-only zero-shot quantization for efficient speech
  recognition,''
\newblock in {\em ICASSP 2022 - 2022 IEEE International Conference on
  Acoustics, Speech and Signal Processing (ICASSP)}, 2022, pp. 4288--4292.

\bibitem{amodei2015deep}
Dario Amodei, Rishita Anubhai, Eric Battenberg, Carl Case, Jared Casper, Bryan
  Catanzaro, Jingdong Chen, Mike Chrzanowski, Adam Coates, Greg Diamos, Erich
  Elsen, Jesse Engel, Linxi Fan, Christopher Fougner, Tony Han, Awni Hannun,
  Billy Jun, Patrick LeGresley, Libby Lin, Sharan Narang, Andrew Ng, Sherjil
  Ozair, Ryan Prenger, Jonathan Raiman, Sanjeev Satheesh, David Seetapun,
  Shubho Sengupta, Yi~Wang, Zhiqian Wang, Chong Wang, Bo~Xiao, Dani Yogatama,
  Jun Zhan, and Zhenyao Zhu,
\newblock ``Deep speech 2: End-to-end speech recognition in english and
  mandarin,'' 2015.

\bibitem{simonyan2014deep}
Karen Simonyan and Andrew Zisserman,
\newblock ``Very deep convolutional networks for large-scale image
  recognition,'' 2014.

\bibitem{zhou2017balanced}
Shuchang Zhou, Yuzhi Wang, He~Wen, Qinyao He, and Yuheng Zou,
\newblock ``Balanced quantization: An effective and efficient approach to
  quantized neural networks,'' 2017.

\bibitem{hannun2014deep}
Awni Hannun, Carl Case, Jared Casper, Bryan Catanzaro, Greg Diamos, Erich
  Elsen, Ryan Prenger, Sanjeev Satheesh, Shubho Sengupta, Adam Coates, and
  Andrew~Y. Ng,
\newblock ``Deep speech: Scaling up end-to-end speech recognition,'' 2014.

\bibitem{43969-speechCommand}
Tara Sainath and Carolina Parada,
\newblock ``Convolutional neural networks for small-footprint keyword
  spotting,''
\newblock in {\em Interspeech}, 2015.

\bibitem{warden2018speech}
Pete Warden,
\newblock ``Speech commands: A dataset for limited-vocabulary speech
  recognition,'' 2018.

\bibitem{Cieri2004TheFC-fisherDataset}
Christopher Cieri, David Miller, and Kevin Walker,
\newblock ``The fisher corpus: a resource for the next generations of
  speech-to-text,''
\newblock in {\em LREC}, 2004.

\bibitem{Panayotov2015LibrispeechAA}
Vassil Panayotov, Guoguo Chen, Daniel Povey, and Sanjeev Khudanpur,
\newblock ``Librispeech: An asr corpus based on public domain audio books,''
\newblock {\em 2015 IEEE International Conference on Acoustics, Speech and
  Signal Processing (ICASSP)}, pp. 5206--5210, 2015.

\bibitem{225858-switchboard}
J.~J. {Godfrey}, E.~C. {Holliman}, and J.~{McDaniel},
\newblock ``Switchboard: telephone speech corpus for research and
  development,''
\newblock in {\em [Proceedings] ICASSP-92: 1992 IEEE International Conference
  on Acoustics, Speech, and Signal Processing}, March 1992, vol.~1, pp.
  517--520 vol.1.

\bibitem{DSQ}
Ruihao Gong, Xianglong Liu, Shenghu Jiang, Tianxiang Li, Peng Hu, Jiazhen Lin,
  Fengwei Yu, and Junjie Yan,
\newblock ``Differentiable soft quantization: Bridging full-precision and
  low-bit neural networks,'' 2019.

\bibitem{IR-Net}
Haotong Qin, Ruihao Gong, Xianglong Liu, Mingzhu Shen, Ziran Wei, Fengwei Yu,
  and Jingkuan Song,
\newblock ``Forward and backward information retention for accurate binary
  neural networks,'' 2019.

\end{thebibliography}

\end{document}